%% file: main.tex
\documentclass[10pt,twocolumn,letterpaper]{article}

\usepackage{cvpr}              
\usepackage[accsupp]{axessibility}  

\input{preamble}
\definecolor{cvprblue}{rgb}{0.21,0.49,0.74}
\usepackage[pagebackref,breaklinks,colorlinks,allcolors=cvprblue]{hyperref}

\def\paperID{} 
\def\confName{CVPR}
\def\confYear{2026}

\title{Three-Step Conditional Diffusion 3D Reconstruction for Light-Field Microscopy\thanks{Accepted to CVPR 2026 Findings.}}
\author{
Qihong Zhao$^{1}$ \quad 
Shaokang Yan$^{1}$ \quad 
Zhimin Qiao$^{1}$ \quad 
Jinjia Wang$^{1}$\thanks{Corresponding authors.} \quad 
Bo Xiong$^{2*}$ \\
$^1$Yanshan University \quad 
$^2$Peking University \\
{\tt\small \{zhaoqh, yskjy, qzm\}@stumail.ysu.edu.cn} \quad 
{\tt\small wjj@ysu.edu.cn} \quad 
{\tt\small xiongbo@pku.edu.cn}
}

\begin{document}
\maketitle
\input{sec/0_abstract}    
\input{sec/1_intro}
\input{sec/2_formatting}

\input{sec/3_finalcopy}

{
    \small
    \bibliographystyle{ieeenat_fullname}
    \bibliography{main}
}


\end{document}

%% file: sec/0_abstract.tex
\begin{abstract}

Light-field microscopy (LFM) enables single-shot capture of multi-angular information from biological samples, supporting real-time volumetric imaging. However, traditional physics-based algorithms often suffer from limited spatial resolution, severe artifacts, and high computational costs. Existing learning-based methods improve inference efficiency but still face limitations in reconstruction accuracy and generalization capability. To address these challenges, this paper proposes a high-fidelity Three-Step Conditional Diffusion (TCD) 3D reconstruction method for LFM. Although conventional diffusion models have achieved remarkable success in generative modeling, their slow sampling process and the inherent trade-off between quality and efficiency hinder their application in real-time 3D imaging. We redesign the diffusion process through a deterministic three-step sampling strategy coupled with a lightweight conditional U-Net, establishing a new paradigm for fast and accurate volumetric reconstruction. Furthermore, an Inter-Class Detection (ICD) module is incorporated to identify out-of-distribution or anomalous inputs during inference, thereby enhancing model stability and reliability. Extensive experiments and cross-dataset evaluations demonstrate that TCD significantly outperforms state-of-the-art methods in both reconstruction fidelity and generalization, providing an efficient and practical 3D reconstruction solution for light-field microscopy.
\end{abstract}

%% file: sec/1_intro.tex
\section{Introduction}
\label{sec:intro}

\begin{figure}[htbp]
  \begin{subfigure}[b]{1\linewidth}
    \includegraphics[width=\linewidth]{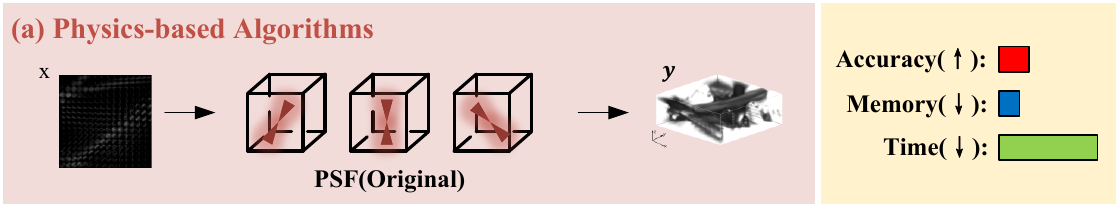}

    \label{f1}
  \end{subfigure}
  \begin{subfigure}[b]{1\linewidth}
    \includegraphics[width=\linewidth]{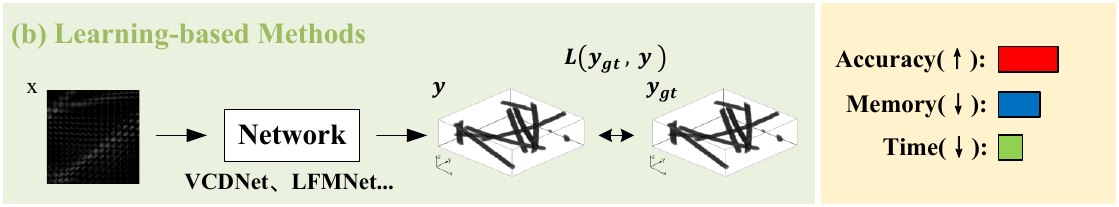}
 
    \label{f1}
  \end{subfigure}
  \begin{subfigure}[b]{1\linewidth}
    \includegraphics[width=\linewidth]{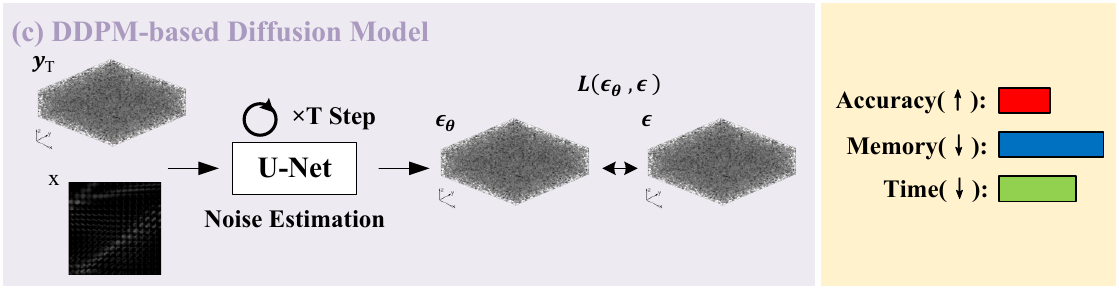}

    \label{f1}
  \end{subfigure}  
  \begin{subfigure}[b]{1\linewidth}
    \includegraphics[width=\linewidth]{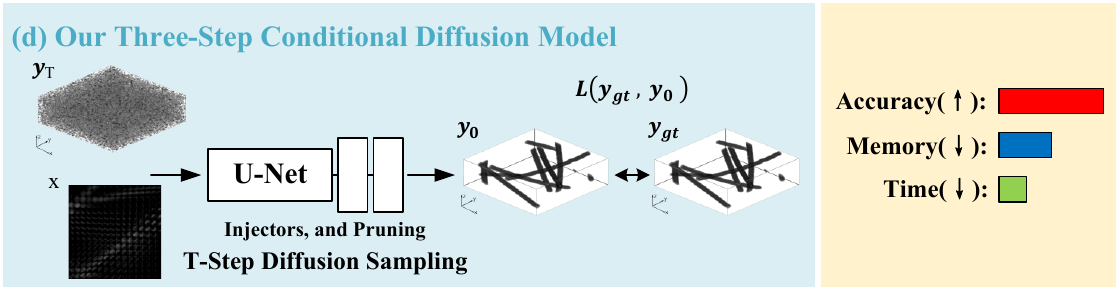}

    \label{f1}
  \end{subfigure}
\caption{\textbf{Comparison of our TCD model with representative imaging methods:} \textbf{(a) Physics-based algorithms:} Based on wave optics, these methods reconstruct 3D volumes by utilizing the system's point spread function (PSF) matrix. \textbf{(b) Learning-based methods:} Deep neural networks employ supervised learning to map light-field images to 3D volumes. \textbf{(c) DDPM-based diffusion model:} This method trains a U-Net-based one-step noise estimator and performs iterative sampling over hundreds of steps using the predicted noise. \textbf{(d) Our improved diffusion model:} This method employs an end-to-end architecture and injects conditional priors into the pruned noise estimation U-Net.}
  \label{fig:single}
\end{figure}

Light-field microscopy (LFM) has emerged as a promising technique for 3D imaging. It enables instantaneous volumetric reconstruction of living specimens and offers unique advantages in capturing dynamic biological processes. These properties make LFM particularly suitable for time-sensitive biological studies \cite{xu20253d,zhou2023aberration,qu2025deep}, such as neural activity imaging, embryonic development monitoring, and cellular dynamics observation \cite{cong2017rapid,yi2023light}.

Traditional reconstruction algorithms based on wave-optics models rely heavily on accurate optical modeling and complex iterative optimization. Despite their strong physical interpretability \cite{broxton2013wave,shroff2013image}, these physics-driven methods often struggle to handle real-world noise and complex biological structures under static priors, resulting in limited spatial resolution, severe reconstruction artifacts, and high computational cost \cite{verinaz2020volume,verinaz2022shift}.

The rapid progress of deep learning has led to the introduction of neural networks in LFM reconstruction. These networks learn the complex nonlinear mapping from light-field images to 3D volumes, achieving notable improvements in both speed and reconstruction quality \cite{li2022incorporating}. Nevertheless, their generalization capability remains limited \cite{zhao2025v2v3d}, which hinders stable performance across diverse experimental conditions and specimen types.

The Denoising Diffusion Probabilistic Model (DDPM), based on Markov chains, has demonstrated exceptional generative capabilities by progressively synthesizing high-fidelity structures \cite{ho2020denoising}. However, its high computational burden and slow inference limit real-time applications in 3D imaging \cite{li2024faster}. To overcome these limitations, we propose a Three-Step Conditional Diffusion (TCD) model for efficient and high-quality light-field 3D reconstruction. Inspired by the deterministic DDIM sampling strategy \cite{liu2024residual,chen2025invertible}, the reverse diffusion process is compressed into just three steps (a number determined experimentally in our framework), achieving an effective trade-off between efficiency and reconstruction quality. Furthermore, the U-Net backbone is redesigned with physics-guided conditional priors injected at each layer and pruned into a compact three-stage architecture, substantially reducing model complexity while preserving reconstruction accuracy.

In addition, we introduce an Inter-Class Detection (ICD) module to evaluate whether a test sample falls within the training distribution. For out-of-distribution (OOD) samples, the system supports rapid fine-tuning with newly collected data to ensure high-quality reconstruction. This mechanism further enhances the model’s generalization and robustness across diverse biological specimens \cite{gwon2023out}, guaranteeing stable and reliable performance under varying experimental conditions.

A comparison of representative 3D reconstruction methods for LFM is shown in Figure~\ref{fig:single}. The main characteristics and contributions of this work are summarized as follows:
\begin{itemize}
    \item \textbf{Efficient diffusion model for LFM:} We are the first to introduce a diffusion-based framework into 3D reconstruction for LFM, employing a deterministic sampling strategy that compresses the generation process from hundreds of iterations to only three, greatly accelerating inference.
    \item \textbf{End-to-end volume prediction:} We reformulate the traditional noise-prediction objective into a direct volume reconstruction paradigm, improving both representational capability and training stability.
    \item \textbf{Light-field guidance and network pruning:} We inject light-field conditional priors and simplify the network structure to reduce model size while preserving reconstruction accuracy.
    \item \textbf{Integration of Inter-Class Detection:} We propose an ICD module that adaptively assesses data distributions to detect out-of-distribution samples and enhance the model’s generalization and robustness across diverse biological specimens.
\end{itemize}

Experimental results on multiple LFM datasets demonstrate that our method achieves superior reconstruction fidelity and strong generalization capability, thus verifying the potential and effectiveness of diffusion models for 3D imaging in biological microscopy.

\section{Related Work}

\subsection{Physics-based 3D reconstruction method}
Traditional confocal fluorescence microscopy relies on sequential scanning to acquire volumetric slices, resulting in long exposure times, photobleaching, and phototoxicity, which limit its application in live-cell dynamic imaging \cite{aimon2019fast,schultz2016advances}. To overcome these constraints, Levoy et al. introduced a microlens array (MLA) into a conventional microscope, pioneering the structure of light-field microscopy and enabling four-dimensional light-field data acquisition \cite{levoy2006light,levoy2009recording}. With its scan-free and high-speed imaging advantages, LFM has been widely applied to real-time volumetric imaging of dynamic biological processes. However, at the microscopic scale, highly transparent specimens cause complex light scattering \cite{levoy2006light,shroff2013image,bishop2011light}, making ray-optics-based models insufficient for accurately describing light propagation and imposing inherent limits on spatial resolution \cite{prevedel2014simultaneous,chen2021three}.

Broxton et al. proposed Richardson–Lucy deconvolution (RLD), a wave-optics-based iterative approach that significantly enhances spatial resolution compared to earlier ray-optics-based models \cite{broxton2013wave,su2023autodeconj,chen2020optical}. Subsequent studies introduced frequency-domain deconvolution and phase-space deconvolution in the angular-frequency domain \cite{stefanoiu2019artifact,stefanoiu20203d,lu2019phase}, both of which improved reconstruction quality and convergence speed. Herman et al. incorporated sparse priors within an alternating direction method of multipliers (ADMM) framework \cite{verinaz2020volume,verinaz2022shift}, while Zhao et al. applied sparse deconvolution to accelerate convergence, suppress noise-induced artifacts, and improve spatial resolution \cite{zhao2022sparse}. Although these methods achieve partial super-resolution, their reliance on sparse priors limits generalizability across diverse microscopic samples.

\subsection{Learning-based 3D reconstruction method}
With the advent of deep learning, convolutional neural networks (CNNs) have been applied to 3D reconstruction for LFM. In 2021, Wang et al. proposed a view-channel-depth network \cite{wang2021real}, achieving high-resolution 3D reconstruction at video frame rates and laying the foundation for subsequent learning-based LFM methods. Subsequent studies introduced advanced architectures, including the 4D convolution-based LFMNet \cite{vizcaino2021learning}, the implicit neural representation network DINER \cite{zhu2024disorder}, and CWFA, which employs invertible normalizing flows with limited training data \cite{page2024fast}. Although these learning-based approaches have achieved significant improvements in reconstruction accuracy and efficiency, they still face challenges in generalization and interpretability \cite{zhao2025v2v3d}.

In recent years, diffusion models have demonstrated remarkable generative capabilities across images, volumes, and videos \cite{chen2024opportunities,dhariwal2021diffusion}. Unlike conventional CNNs that directly map inputs to outputs, diffusion models learn data distributions by estimating noise, enabling them to capture complex spatial structures and fine-grained details. Their iterative denoising process enhances robustness and generalization, while the forward–reverse diffusion can be formulated as a Markov chain, providing physical interpretability \cite{liu2024residual}. By incorporating conditional information as guidance \cite{dufour2024don}, conditional diffusion models achieve controllable and semantically consistent generation \cite{ni2023conditional,zhang2023adding}. For instance, in text-to-image diffusion \cite{saharia2022photorealistic,rombach2022high}, textual conditioning enables fine-grained control over spatial and semantic attributes, resulting in high-resolution, realistic outputs. Motivated by these properties, diffusion models offer a theoretical foundation for high-fidelity, controllable, and interpretable reconstruction in complex tasks such as 3D reconstruction for LFM \cite{ni2023conditional,ruan2023mm}.

\subsection{Inter-Class Detection for LFM}
In high-fidelity LFM 3D reconstruction, distributional variations in input samples challenge model generalization, particularly when encountering novel biological specimens and experimental conditions. To address this issue, inter-class detection modules have been proposed to estimate the distributional confidence of input data. Early out-of-distribution detection methods mainly relied on simple confidence metrics or softmax outputs \cite{lee2018simple}. However, these approaches often perform unstably in complex, high-dimensional microscopy data. Subsequently, methods based on feature-space distances or generative models were developed to more accurately determine whether a sample belongs to the training distribution. For instance, ODIN \cite{liang2017enhancing} enhances OOD recognition using temperature scaling and input perturbations, while the Mahalanobis distance-based approach \cite{yang2020out} estimates inter-class confidence from feature distribution statistics. Although these methods achieve notable performance in general computer vision tasks, their direct application to LFM reconstruction remains challenging due to the high dimensionality of input light-field data and complex optical conditions \cite{vasiliauskaite2024generalization}.

%% file: sec/2_formatting.tex
\section{Method}
In this section, we first introduce the traditional DDPM framework applied to 3D reconstruction for LFM. Next, we describe TCD based on this framework, with a detailed presentation of its core design components. Finally, we present the ICD module, forming the complete reconstruction pipeline.

\subsection{Preliminaries: Diffusion Models}

\begin{figure*}[htbp]
    \centering
    \includegraphics[width=1\linewidth]{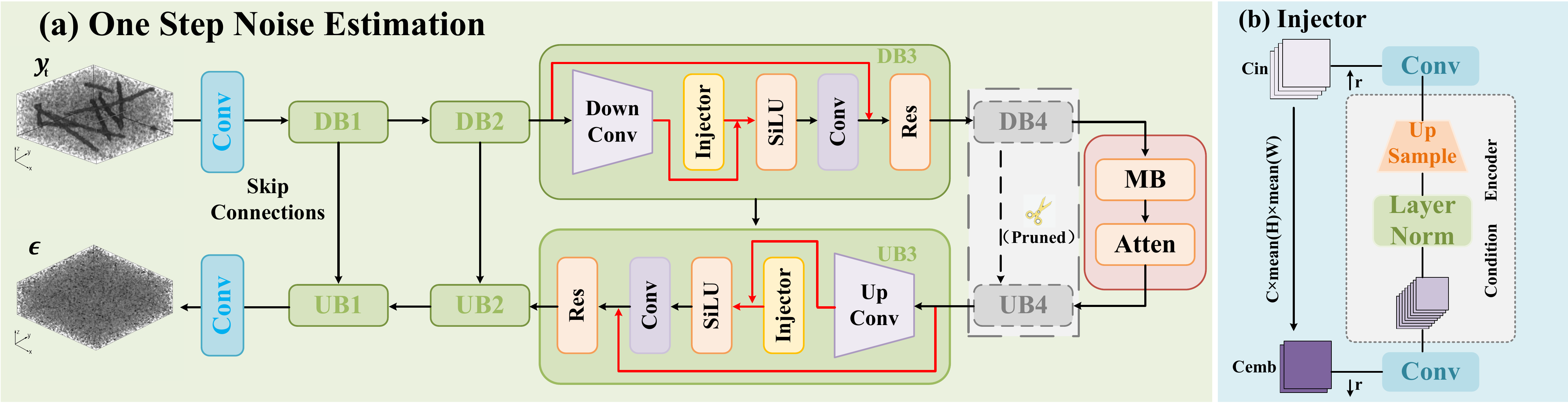}
    \caption{\textbf{Overview of the U-Net-based one-step noise estimation framework.} (a) The main U-Net architecture, composed of a sequence of Downsampling Blocks (DBs), Upsampling Blocks (UBs), and a Middle Block (MB) that includes a self-attention (Atten) module. Injectors are integrated within these blocks to enhance reconstruction fidelity. To reduce model complexity, select blocks are pruned. (b) The Injector module. This module is designed to fuse the prior guidance provided by the light-field input \( C_{\text{in}} \). It uses a specific network structure combined with residual connections to inject physics information into the Sampling Blocks of the U-Net.}
    \label{fig3}
\end{figure*}

Our framework is grounded in the DDPM paradigm, which formulates data generation as a gradual denoising process driven by a Markov chain. Specifically, the modeling framework integrates thermodynamic diffusion theory with deep learning-based probabilistic inference \cite{sohl2015deep}, leveraging the physical interpretability of diffusion processes and the representational power of neural networks. It consists of two main stages: a forward diffusion process and a reverse denoising process. To adapt the DDPM for LFM 3D reconstruction, we integrate light-field images and noisy volumes as inputs, enabling effective data fusion for volumetric reconstruction.

\subsubsection{Forward Diffusion}

Given a clean 3D volume \(\mathbf{y}_0 \in \mathbb{R}^{D \times H \times W}\), Gaussian noise is gradually added through a Markov process over \(T\) timesteps to generate a pure noise volume \(\mathbf{y}_T\). At each timestep \( t \), the process is defined as:
\begin{equation}
\mathbf{y}_t = \sqrt{\alpha_t} \mathbf{y}_{t-1} + \sqrt{1 - \alpha_t} \cdot \boldsymbol{\epsilon},
\end{equation}
where \(\boldsymbol{\epsilon} \sim \mathcal{N}(0, \mathbf{I})\) is standard Gaussian noise, \(\mathbf{y}_t\) denotes the noisy volume at timestep \(t\), and \(\alpha_t = 1 - \beta_t\) with the noise schedule \(\beta_t \in [0, 1]\), \(\beta_0 = 0\).

According to the variance-preserving stochastic differential equation (VP-SDE) \cite{song2020score}, the noisy volume \(\mathbf{y}_t\) can also be directly computed from the original clean volume \(\mathbf{y}_0\) as:
\begin{equation}
\mathbf{y}_t = \sqrt{\bar{\alpha}_t} \mathbf{y}_0 + \sqrt{1 - \bar{\alpha}_t} \cdot \boldsymbol{\epsilon},
\end{equation}
where \(\bar{\alpha}_t = \prod_{i=1}^{t} \alpha_i\) is the cumulative product of \(\alpha_i\).

During training, a random timestep \(t \sim \mathcal{U}(0, T)\) is selected. The noisy volume \(\mathbf{y}_t\), the corresponding conditional light-field image \(\mathbf{x}\), and the timestep \(t\) are fed into a U-Net-based noise prediction network \(\boldsymbol{\epsilon}_\mathbf{\theta}\) (where \(\mathbf{\theta}\) denotes the trainable parameters of the diffusion network). The model is optimized by minimizing the mean squared error (MSE) between the true noise \(\boldsymbol{\epsilon}\) and the predicted noise \(\boldsymbol{\epsilon}_\mathbf{\theta}\):
\begin{equation}
\arg\min_{\mathbf{\theta}} \left\| \boldsymbol{\epsilon} - \boldsymbol{\epsilon}_\mathbf{\theta}\left( \sqrt{\bar{\alpha}_t} \mathbf{y}_0 + \sqrt{1 - \bar{\alpha}_t} \cdot \boldsymbol{\epsilon},\ \mathbf{x},\ t \right) \right\|_2^2.
\end{equation}

\subsubsection{Reverse Denoising}

To recover the clean 3D volume, we adapt the reverse diffusion process, which iteratively approximates the posterior distribution $p_\mathbf{\theta}(\mathbf{y}_{t-1} \mid \mathbf{y}_t)$ based on Bayes' theorem and Gaussian combination rules \cite{ho2020denoising}:
\begin{equation}
p_\mathbf{\theta}(\mathbf{y}_{t-1} \mid \mathbf{y}_t) = \mathcal{N}(\mathbf{y}_{t-1}; \mathbf{\mu}_\mathbf{\theta}(\mathbf{y}_t, \mathbf{x}, t), \sigma_t^2 \mathbf{I}),
\end{equation}
where \(\mathbf{\mu}_\mathbf{\theta}(\mathbf{y}_t, \mathbf{x}, t)\) and \(\sigma_t\) denote the predicted mean and the sampling variance, respectively, at each timestep \(t\).

Instead of modeling the full data distribution, our network learns to predict the noise, enabling the following update rule:
\begin{equation}
\mathbf{y}_{t-1} = \frac{1}{\sqrt{\alpha_t}} \left( \mathbf{y}_t - \frac{\beta_t}{\sqrt{1 - \bar{\alpha}_t}} \cdot \boldsymbol{\epsilon}_\mathbf{\theta}(\mathbf{y}_t, \mathbf{x}, t) \right) + \sigma_t \cdot \boldsymbol{\epsilon}_{\text{rand}},
\end{equation}
where \(\boldsymbol{\epsilon}_\mathbf{\theta}(\mathbf{y}_t, \mathbf{x}, t)\) is the predicted noise at step \(t\), and \(\boldsymbol{\epsilon}_{\text{rand}} \sim \mathcal{N}(0, \mathbf{I})\) is a randomly sampled Gaussian noise introduced to maintain diversity.

This denoising process proceeds in reverse from $t = T$ to $t = 0$, gradually removing the accumulated noise and producing the final clean volume $\mathbf{y}_0$ conditioned on the light-field measurement $\mathbf{x}$. A visualization of the DDPM denoising process over 500 steps is provided in the supplementary material.


\begin{figure*}[htbp]
    \centering
    \includegraphics[width=1\linewidth]{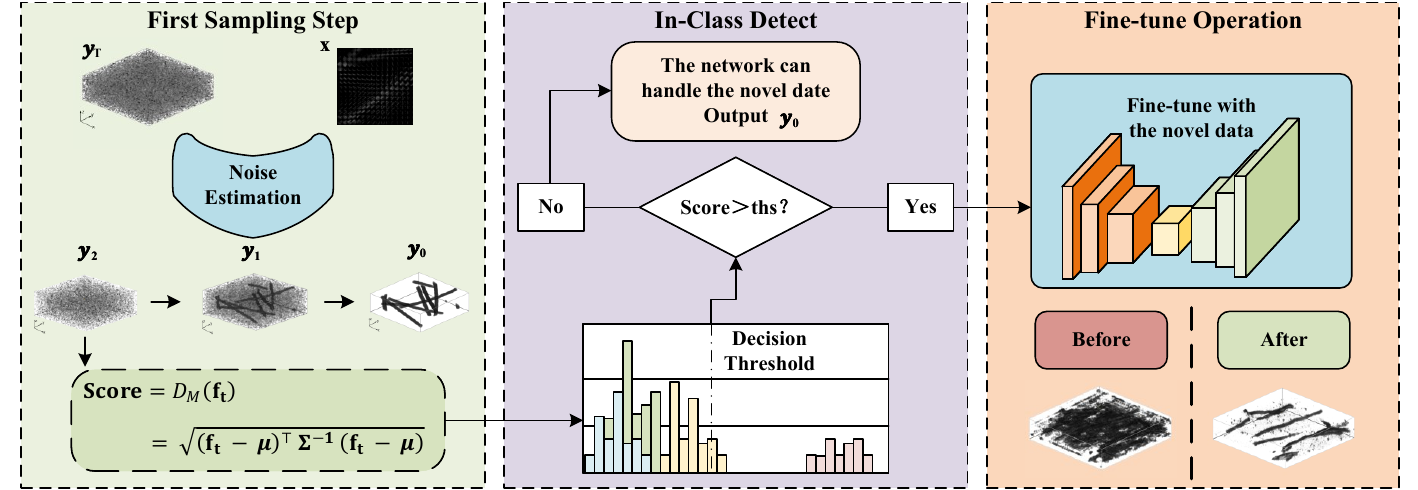}
    \caption{\textbf{A schematic overview of the complete workflow incorporating ICD detection.} The figure illustrates how out-of-distribution samples are identified and subsequently handled.}
    \label{fig4}
\end{figure*}

\subsection{Three-Step Conditional Diffusion}

To address the inefficiency and stochasticity of conventional DDPMs, we propose a deterministic Three-Step Conditional Diffusion framework for LFM 3D reconstruction. TCD achieves high-fidelity reconstruction while significantly accelerating inference through a three-step deterministic sampling strategy. By integrating an end-to-end optimized, physics-guided lightweight U-Net, our framework effectively balances reconstruction quality and computational cost.

\subsubsection{Deterministic Sampling Strategy (T=3)}
Standard DDPM inference requires hundreds of reverse denoising steps, resulting in high computational cost and slow inference speed. To improve efficiency, we adopt a DDIM-inspired deterministic sampling strategy, which replaces stochastic noise sampling with a deterministic update process using fixed timesteps, thereby significantly accelerating the inference process.

During denoising, the noise predictor \(\boldsymbol{\epsilon}_\mathbf{\theta}\) guides the denoising direction. The deterministic sampling involves two key steps:
\begin{equation}
\begin{aligned}
\mathbf{y}_{0|t} &= \frac{1}{\sqrt{\bar{\alpha}_t}} \left( \mathbf{y}_t - \sqrt{1 - \bar{\alpha}_t} \cdot \boldsymbol{\epsilon}_\mathbf{\theta}(\mathbf{y}_t, \mathbf{x}, t) \right), \\
\mathbf{y}_{t-1} &= \sqrt{\bar{\alpha}_{t-1}} \cdot \mathbf{y}_{0|t} + \sqrt{1 - \bar{\alpha}_{t-1}} \cdot \boldsymbol{\epsilon}_\mathbf{\theta}(\mathbf{y}_t, \mathbf{x}, t).
\end{aligned}
\end{equation}

This process can be equivalently combined as:
\begin{equation}
\begin{aligned}
\mathbf{y}_{t-1} &= \sqrt{\bar{\alpha}_{t-1}} \left( \frac{\mathbf{y}_t - \sqrt{1 - \bar{\alpha}_t} \cdot \boldsymbol{\epsilon}_\mathbf{\theta}(\mathbf{y}_t, \mathbf{x}, t)}{\sqrt{\bar{\alpha}_t}} \right) \\
&\quad + \sqrt{1 - \bar{\alpha}_{t-1}} \cdot \boldsymbol{\epsilon}_\mathbf{\theta}(\mathbf{y}_t, \mathbf{x}, t).
\end{aligned}
\end{equation}

Unlike DDPM and DDIM, our sampling strategy constructs a deterministic equivalent reverse process, bypassing the full Markov chain by reformulating sampling as a 3D-optimized unrolled network. This formulation enables high-fidelity inference with significantly fewer steps. Furthermore, flexible timestep scheduling allows for a controllable trade-off between efficiency and quality, catering to real-time 3D reconstruction and generation. As validated in Figure~\ref{fig:step_analysis}, TCD uses only three deterministic steps to achieve over a 90\% reduction in sampling cost while maintaining reconstruction accuracy.

\subsubsection{End-to-End Reconstruction Network}
To enable a direct, end-to-end reconstruction mapping from an initial noise volume to the ground truth 3D volume \(\mathbf{y}_{\mathrm{gt}}\), we extend the deterministic sampling acceleration strategy into a full sampling network. This network is composed of \(T\) sequentially stacked U-Net modules, formulated as:
\begin{equation}
F = F_1 \circ F_2 \circ \cdots \circ F_T, \quad \mathbf{y}_{t-1} = F_t(\mathbf{y}_t, \mathbf{x}, t),
\end{equation}
where each function \( F_t(\mathbf{y}_t, \mathbf{x}, t) \) corresponds to a full sampling step from \(\mathbf{y}_t\) to \(\mathbf{y}_{t-1}\), and \( \circ \) denotes the composition of sequential steps.

While conventional DDPM frameworks train only the noise estimator \(\boldsymbol{\epsilon}_{\mathbf{\theta}}\), our model optimizes the entire sampling trajectory by jointly minimizing the reconstruction loss between the final predicted volume and ground truth volumes:
\begin{equation}
\mathcal{L}_{\mathrm{recon}} = \left\| \mathbf{y}_{\mathrm{gt}} - \mathbf{y}_0 \right\|_2^2,
\label{eq:recon_loss}
\end{equation}
By constructing an end-to-end network, the model can more effectively learn the final reconstruction objective, thereby improving adaptability and performance in high-precision 3D reconstruction tasks.

This unified design bridges probabilistic diffusion with supervised learning, enabling an efficient and task-oriented optimization pipeline that enhances both reconstruction fidelity and computational efficiency.

\begin{table*}[t]
\caption{\textbf{Quantitative comparison with state-of-the-art (SOTA) methods across five representative biological categories.}
Bold indicates the best and underline indicates the second-best results.}
\label{tab:bio_results_exclude_b1}
\centering
\scriptsize
\setlength{\tabcolsep}{7pt}
\begin{tabular}{l|cc|cc|cc|cc|cc|cc}
\toprule
\textbf{Method} & 
\multicolumn{2}{c|}{\textbf{RLD}} & 
\multicolumn{2}{c|}{\textbf{VCDNet}} & 
\multicolumn{2}{c|}{\textbf{LFMNet}} & 
\multicolumn{2}{c|}{\textbf{CWFA}} & 
\multicolumn{2}{c|}{\textbf{DDPM}} & 
\multicolumn{2}{c}{\textbf{Ours (TCD)}} \\
\midrule
\textbf{Scene} & PSNR $\uparrow$ & SSIM $\uparrow$ & PSNR $\uparrow$ & SSIM $\uparrow$ & PSNR $\uparrow$ & SSIM $\uparrow$ & PSNR $\uparrow$ & SSIM $\uparrow$ & PSNR $\uparrow$ & SSIM $\uparrow$ & PSNR $\uparrow$ & SSIM $\uparrow$ \\
\midrule
Tubulin& 27.60 & 0.749 & 35.58 & \underline{0.981} & \underline{36.28} & 0.979 & 34.44 & 0.686 & 34.32 & 0.966 & \textbf{38.41}& \textbf{0.985} \\
Vessel& 28.88 & 0.662 & 35.15 & 0.800 & 35.56 & \underline{0.814} & \underline{35.86} & 0.806 & 30.26 & 0.409 & \textbf{36.04} & \textbf{0.825} \\
Bcell& 37.02 & 0.773 & 40.60 & 0.786 & \underline{47.49} & \underline{0.972} & 33.63 & 0.629 & 41.03 & 0.875 & \textbf{47.54} & \textbf{0.981} \\
Mito& 42.05 & 0.926 & 35.96 & 0.643 & \underline{44.99} & \underline{0.949} & 33.19 & 0.679 & 38.20 & 0.889 & \textbf{45.77} & \textbf{0.957} \\
Dendrite& 35.59 & 0.724 & 35.58 & 0.570 & 37.60 & \underline{0.866} & 30.87 & 0.735 & \underline{37.79} & 0.808 & \textbf{38.85} & \textbf{0.872} \\
\bottomrule
\end{tabular}
\end{table*}

\begin{figure*}[htbp]
    \centering
    \includegraphics[width=1\linewidth]{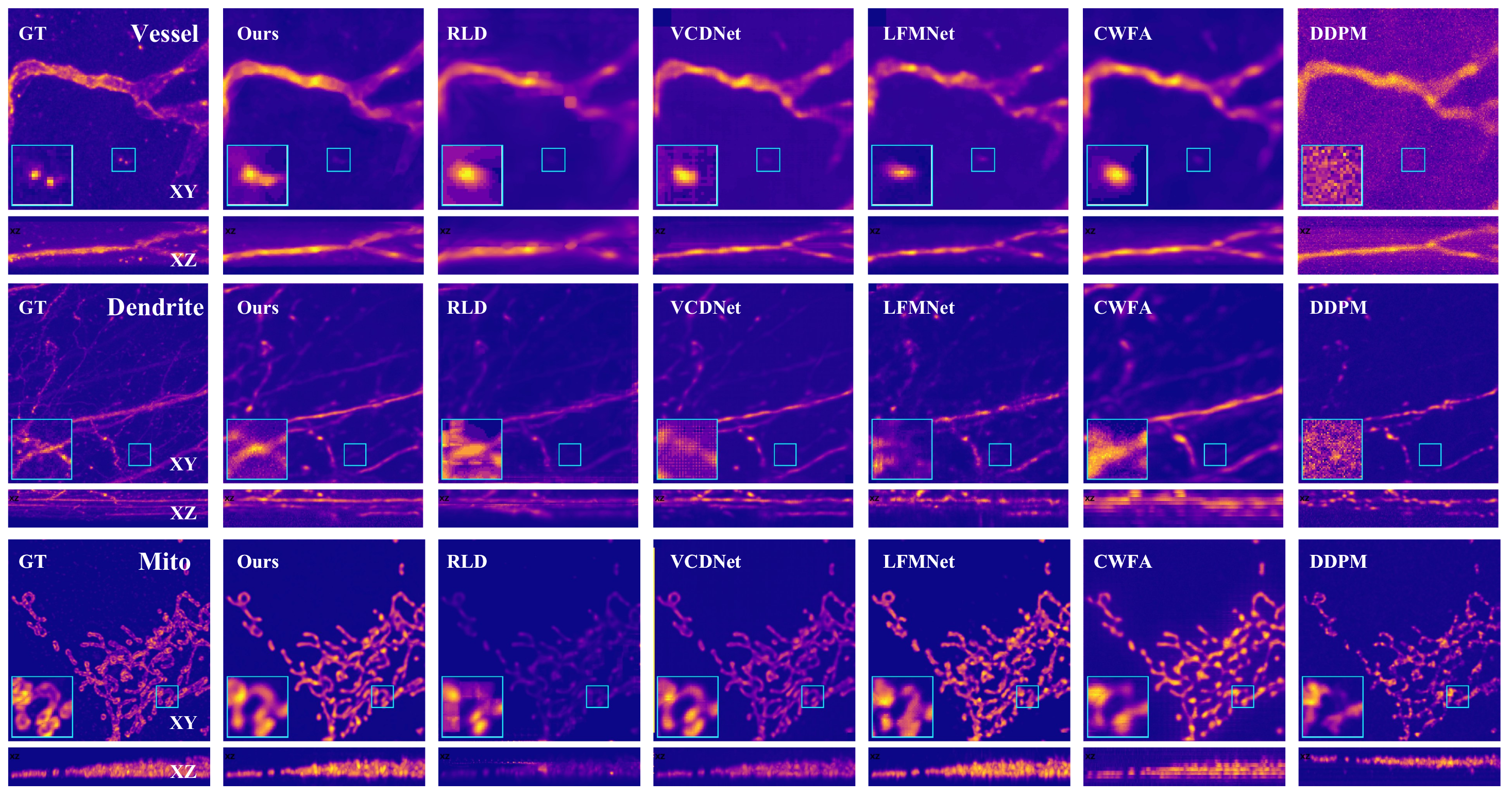}
    \caption{\textbf{Qualitative comparisons on the Vessel, Dendrite, and Mito datasets.}}
    \label{fig5}
\end{figure*}

\subsubsection{Light-Field-Guided and Lightweight Design}

In the TCD model, sequential stacking of multiple U-Net modules causes the network capacity to grow linearly with the number of modules, leading to substantial memory usage and computational overhead. To address this issue, we prune the U-Net backbone, reducing the conventional four-stage architecture to a three-stage design, which significantly reduces the number of parameters and computational cost. However, this reduction in network parameters can weaken its representational capacity, potentially leading to a decrease in reconstruction fidelity.

To compensate for this potential loss and preserve reconstruction fidelity, we introduce a physics-guided conditional prior through a compact Conditional Information Injector. Specifically, the Condition Encoder extracts multi-scale features from the light-field input $\mathbf{x}$, which are progressively upsampled, normalized, and averaged to form the condition embedding $\mathbf{c}_{\text{emb}}$. This embedding is then fused into the hidden feature map $\mathbf{h}$ through a lightweight injection operation:
\begin{equation}
\mathbf{y}_{\text{cond}} = \operatorname{SiLU} \big( \operatorname{Conv}(\mathbf{h}) + \mathbf{c}_{\text{emb}} \big),
\end{equation}
where $\mathbf{y}_{\text{cond}}$ denotes the fused output.

This design introduces only approximately 0.6 M of additional parameters, thereby compensating for the impact of network pruning. By integrating network pruning with conditional feature injection, we establish an efficient, physics-aware denoising framework that effectively balances reconstruction fidelity and computational cost, as illustrated in Figure~\ref{fig3} and Table~\ref{tab:ablation}.

\begin{table*}[t]
\caption{\textbf{Quantitative comparison between models trained on a single dataset (left) and a mixed dataset (right).}}
\label{tab:dendrite_mixed_compare}
\centering
\scriptsize
\setlength{\tabcolsep}{3pt}
\begin{tabular}{l|cc|cc|cc|cc||cc|cc|cc|cc}
\toprule
{\textbf{Method}}& \multicolumn{8}{c||}{\textbf{Train: Dendrite Dataset}} 
& \multicolumn{8}{c}{\textbf{Train: Mixed Dataset}} \\
\cmidrule(lr){2-9} \cmidrule(lr){10-17}
& \multicolumn{2}{c|}{\textbf{Ours (TCD)}} 
& \multicolumn{2}{c|}{\textbf{VCDNet}} 
& \multicolumn{2}{c|}{\textbf{LFMNet}} 
& \multicolumn{2}{c||}{\textbf{DDPM}} 
& \multicolumn{2}{c|}{\textbf{Ours (TCD)}} 
& \multicolumn{2}{c|}{\textbf{VCDNet}} 
& \multicolumn{2}{c|}{\textbf{LFMNet}} 
& \multicolumn{2}{c}{\textbf{DDPM}} \\
\midrule
{\textbf{Scene}}& PSNR $\uparrow$ & SSIM $\uparrow$ & PSNR $\uparrow$ & SSIM $\uparrow$ & PSNR $\uparrow$ & SSIM $\uparrow$ & PSNR $\uparrow$ & SSIM $\uparrow$
& PSNR $\uparrow$ & SSIM $\uparrow$ & PSNR $\uparrow$ & SSIM $\uparrow$ & PSNR $\uparrow$ & SSIM $\uparrow$ & PSNR $\uparrow$ & SSIM $\uparrow$ \\
\midrule
Bcell     & \textbf{37.15}& \textbf{0.885}& 34.41& 0.693& \underline{37.02}& \underline{0.841}& 32.99& 0.751& \textbf{39.822} & \textbf{0.926} & 37.647 & 0.820 & \underline{39.368} & \underline{0.922} & 33.061 & 0.758 \\
Dendrite  & \textbf{38.85}& \textbf{0.872}& 35.58& 0.570& 37.60& \underline{0.866}& \underline{37.79}& 0.808& \textbf{37.287} & \textbf{0.834} & \underline{34.629} & \underline{0.773} & 34.455 & 0.730 & 33.211 & 0.654 \\
Mito      & \underline{34.51}& \underline{0.816}& 31.87& 0.316& \textbf{36.95}& \textbf{0.899}& 31.25& 0.583& \textbf{41.346} & \textbf{0.935} & 35.883 & 0.649 & \underline{38.459} & \underline{0.916} & 32.521 & 0.633 \\
Vessel    & \underline{31.43}& \underline{0.647}& \textbf{32.91}& \textbf{0.657}& 29.75& 0.642& 28.98& 0.479& \textbf{33.230} & \textbf{0.728} & 31.734 & 0.589 & \underline{32.672} & \underline{0.717} & 29.507 & 0.500 \\
Tubulin   & 27.69& \underline{0.640}& \textbf{30.54}& 0.354& \underline{28.49}& \textbf{0.794}& 26.86& 0.523& \textbf{37.861} & \textbf{0.965} & 33.977 & 0.690 & \underline{34.542} & \underline{0.938} & 30.098 & 0.631 \\
\bottomrule
\end{tabular}
\end{table*}

\begin{table}[htbp]
\centering
\caption{\textbf{Efficiency comparison with SOTA methods.}}
\label{tab:method_comparison1}
\footnotesize
\setlength{\tabcolsep}{2pt}
\begin{tabular}{lccccc}
\toprule
Method & PSNR $\uparrow$ & SSIM $\uparrow$ &Train. (h)& Param. (M)& Inf. (s)\\
\midrule
RLD     & 34.63& 0.767&---   & ---   & 194.991 \\
VCDNet  & 36.17& 0.756&2.0& 18.8& 0.042 \\
LFMNet  & 40.38& 0.916&2.0& 22.0& 0.021\\
CWFA    & 33.20& 0.707&2.6& 43.5& 0.214 \\
DDPM    & 36.32& 0.789&2.2& 37.9& 4.256 \\
Ours (TCD) & 41.72& 0.924&2.6& 32.5& 0.062\\
\bottomrule
\end{tabular}
\end{table}

\subsection{Inter-Class Detection Module}
Learning-based LFM volumetric reconstruction methods generally assume that test samples follow a distribution similar to that of the training data. However, in practical microscopy scenarios, variations in imaging conditions—such as optical settings, noise levels, or biological sample types—can significantly degrade generalization and result in unstable reconstruction quality. To address this issue, we propose an Inter-Class Detection module for reliable distribution estimation and adaptive inference during testing.

The ICD module leverages the volume $\mathbf{y}_{t-1}$ obtained from the first denoising step of the TCD framework. This intermediate volume, which approximately follows a Gaussian distribution and contains reconstructed structural information, is well-suited for distribution modeling. For each training sample, we compute statistical descriptors of the 3D volume, including mean, standard deviation, variance, and L1/L2 norms. These features are aggregated across all training samples and modeled as a multivariate Gaussian distribution to represent the inter-class feature space:

\begin{equation}
p(\mathbf{f}) = \mathcal{N}(\boldsymbol{\mu}, \boldsymbol{\Sigma}),
\end{equation}
where $\boldsymbol{\mu}$ and $\boldsymbol{\Sigma}$ denote the mean vector and covariance matrix of inter-class features, respectively. These parameters are fixed after training and serve as a baseline for anomaly detection during inference.

During testing, $\mathbf{y}_{t-1}$ is extracted for each input sample, and its Mahalanobis distance to the in-distribution (ID) is computed \cite{yang2020out}:

\begin{equation}
\text{Score} = D_M(\mathbf{f}_t) = \sqrt{(\mathbf{f}_t - \boldsymbol{\mu})^\top \boldsymbol{\Sigma}^{-1} (\mathbf{f}_t - \boldsymbol{\mu})},
\end{equation}
where $\mathbf{f}_t$ denotes the $t$-th feature vector. Detection threshold ($\text{ths}$) is determined from the Mahalanobis distance distribution of training samples (e.g., the 85th percentile). Test samples with $\text{Score} < \text{ths}$ are considered ID and undergo standard TCD reconstruction. Otherwise, they are flagged as potential OOD samples, triggering adaptive strategies such as adding the sample to the training set for fine-tuning or rejecting unreliable outputs.

Experiments show that the module enhances stability for OOD samples, enabling robust volumetric reconstruction across diverse biological specimens.

%% file: sec/3_finalcopy.tex
\section{Experiments}

In this section, we first introduce the experimental setup and datasets for evaluation. Next, we present both quantitative and qualitative comparisons with other SOTA methods. Then, we conduct ablation studies to examine the various components of TCD. Finally, we introduce the ICD module for performance evaluation. For additional experiments, please refer to the supplementary materials.

\subsection{Experimental Setup and Datasets}
We evaluated the performance of the proposed method using both synthetic and real-world biological volumetric data. The experimental dataset consists of five types of 3D volumes, including Tubulin, Vessel, Mito, Dendrite, and Bead cell (Bcell). First, the original volumetric data were cropped and resampled to achieve a uniform resolution and remove redundant edge information, yielding standardized 3D volumes. Subsequently, these volumes were projected into light-field images using a physics-based forward projection method, simulating the acquisition of light-field information in practical microscopy. To ensure representative testing, we selected five typical cases from the dataset as the test set, covering different biological structures.
 
All experiments were conducted on an NVIDIA RTX 3090 GPU to handle the computational demands of large-scale volumetric data. In each experiment, reconstruction quality was quantitatively evaluated using metrics such as PSNR and SSIM, validating the performance of all methods across diverse biological structures and imaging conditions.

\subsection{Comparative Experiments}

We performed a comprehensive quantitative comparison between the proposed method and several representative state-of-the-art LFM 3D reconstruction approaches, including the traditional algorithm-based Richardson–Lucy deconvolution (RLD, 8 iterations) \cite{broxton2013wave}, the convolutional neural network-based VCDNet \cite{wang2021real}, the four-dimensional convolution model LFMNet \cite{vizcaino2021learning}, the invertible wavelet flow model (CWFA) \cite{page2024fast}, and our baseline diffusion-based model (DDPM) \cite{ho2020denoising}.  
All methods were trained and evaluated under identical experimental settings to ensure a fair comparison.  

As shown in Table~\ref{tab:bio_results_exclude_b1}, we report the quantitative metrics for each model trained and tested on its corresponding dataset. The reconstructed results and the ground-truth 3D volumes (GT) in the XY and XZ central planes are further illustrated in Figure~\ref{fig5}. The results demonstrate that the proposed TCD framework consistently achieves the best performance across all tested scenes, significantly outperforming other methods in detail restoration and exhibiting superior spatial fidelity.

The cross-sample evaluation results obtained when models are trained on the dendrite dataset (left), as well as the overall performance on a mixed dataset comprising five biological categories (right), are presented in Table~\ref{tab:dendrite_mixed_compare}. The results show that TCD achieves optimal or sub-optimal performance in cross-sample testing under single-sample training conditions, and significantly outperforms other methods when trained and tested on multi-category samples. These findings demonstrate the strong generalization and robustness capability of TCD. Furthermore, the supplementary materials provide visual comparisons of reconstruction results between TCD and other methods after training on the mixed dataset, further demonstrating its superiority in detail and structural reconstruction.

A comprehensive comparison of all methods in terms of model complexity and computational efficiency is provided in Table~\ref{tab:method_comparison1}, including PSNR, SSIM (averaged results from Table~\ref{tab:bio_results_exclude_b1}), training time (↓), parameter count (↓), and inference time (↓). The results show that TCD offers significant advantages in computational efficiency compared with traditional algorithm-based approaches and the DDPM model.

\subsection{Ablation Experiments} 

\begin{figure}[t]
   \centering
    \includegraphics[width=1\linewidth]{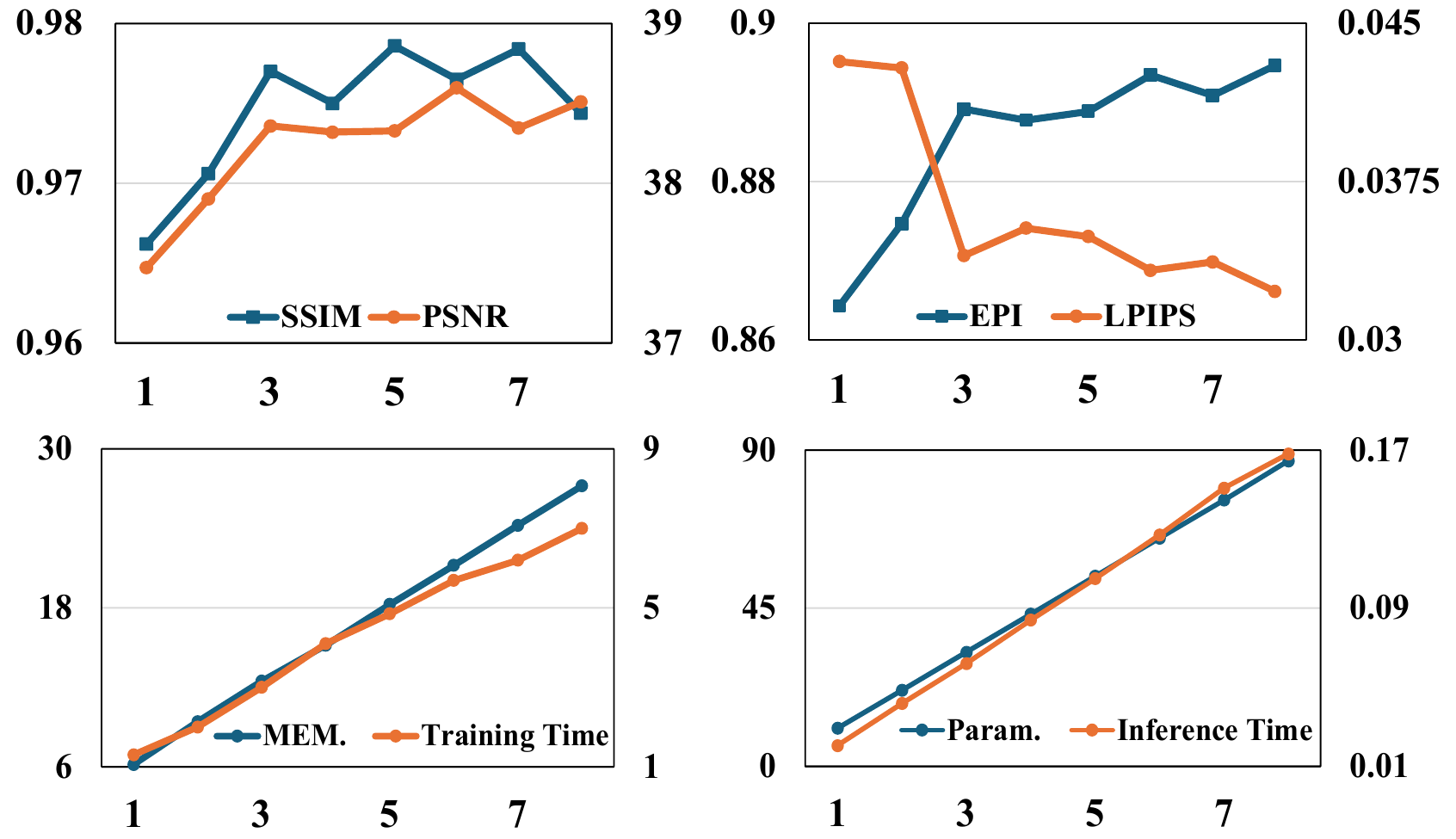}

  \caption{\textbf{Comparison of TCD model variants with varying step \(1 \leq T \leq 8\).}}
  \label{fig:step_analysis}
\end{figure}

The ablation experiments were conducted on the Tubulin dataset. As shown in Figure~\ref{fig:step_analysis}, we evaluated the impact of the sampling step \(T\) on reconstruction quality and computational cost, including PSNR (↑), SSIM (↑), LPIPS (↓, learned perceptual image patch similarity), EPI (↑, edge preservation index), training time (↓, h), GPU memory consumption (↓, GB), parameter count (↓, M), and inference time (↓, s). The results indicate that when \(T < 3\), the reconstruction performance increases significantly, while for \(T > 3\), the performance gain tends to saturate. Meanwhile, the computational cost grows approximately linearly. Therefore, we select \(T = 3\) as the optimal trade-off between quality and computational efficiency.

As shown in Table~\ref{tab:ablation}, we conducted systematic ablation studies on the key components of the TCD framework, including the sampling steps \(T\), end-to-end training (E2E), conditional information injector (Inj.), and U-Net pruning (Pru.), to investigate their effects on model performance. The evaluation metrics include PSNR (↑), SSIM (↑), EPI (↑), Inf. (↓, s), and Param. (↓, M). The experimental results demonstrate that the end-to-end TCD significantly outperforms DDPM and DDIM in reconstruction quality, while reducing the required sampling steps by over 90\% compared to DDPM. This reduction leads to a two-order-of-magnitude acceleration in inference speed. Although pruning the U-Net slightly decreases reconstruction accuracy, it reduces model parameters by approximately 70\%, substantially lowering GPU memory consumption. Moreover, introducing the conditional information injector compensates for the performance degradation caused by pruning, achieving an optimal balance between accuracy and efficiency.

\begin{table}[t]
\centering
\scriptsize
\setlength{\tabcolsep}{3pt}  
\caption{\textbf{Ablation study of TCD under different configurations.} }
\label{tab:ablation}
\begin{tabular}{lcccccccccc}
\toprule
Method & $T$ & E2E & Inj. & Pru. & PSNR & SSIM & EPI & LPIPS & Inf. & Param. \\
\midrule
DDPM & 300 & $\times$ & $\times$ & $\times$ & 33.02 & 0.954 & 0.617 & 0.123 & 3.080 & 37.9 \\
DDPM & 500 & $\times$ & $\times$ & $\times$ & 33.65 & 0.961 & 0.673 & 0.097 & 5.101 & 37.9 \\
DDPM & 700 & $\times$ & $\times$ & $\times$ & 33.24 & 0.958 & 0.607 & 0.117 & 6.960 & 37.9 \\
DDIM & 50  & $\times$ & $\times$ & $\times$ & 33.17 & 0.957 & 0.649 & 0.104 & 0.552 & 37.9 \\
Ours & 3 & $\checkmark$ & $\times$ & $\times$ & 37.86 & \underline{0.971} & 0.879 & 0.040 & 0.076 & 114.5 \\
Ours & 3 & $\checkmark$ & $\times$ & $\checkmark$ & 37.64 & 0.963 & 0.865 & 0.047 & \textbf{0.062} & \textbf{31.9} \\
Ours & 3 & $\checkmark$ & $\checkmark$ & $\times$ & \textbf{38.46} & \textbf{0.985} & \textbf{0.893} & \textbf{0.032} & \underline{0.075} & 115.7 \\
Ours & 3 & $\checkmark$ & $\checkmark$ & $\checkmark$ & \underline{38.41} & \textbf{0.985} & \underline{0.892} & \textbf{0.032} & \textbf{0.062} & \underline{32.5} \\
\bottomrule
\end{tabular}
\end{table}



\subsection{Inter-Class Detection for Sample Adaptation}

\begin{figure}[t]
  \centering
  \begin{subfigure}{1\columnwidth}
    \includegraphics[width=\linewidth]{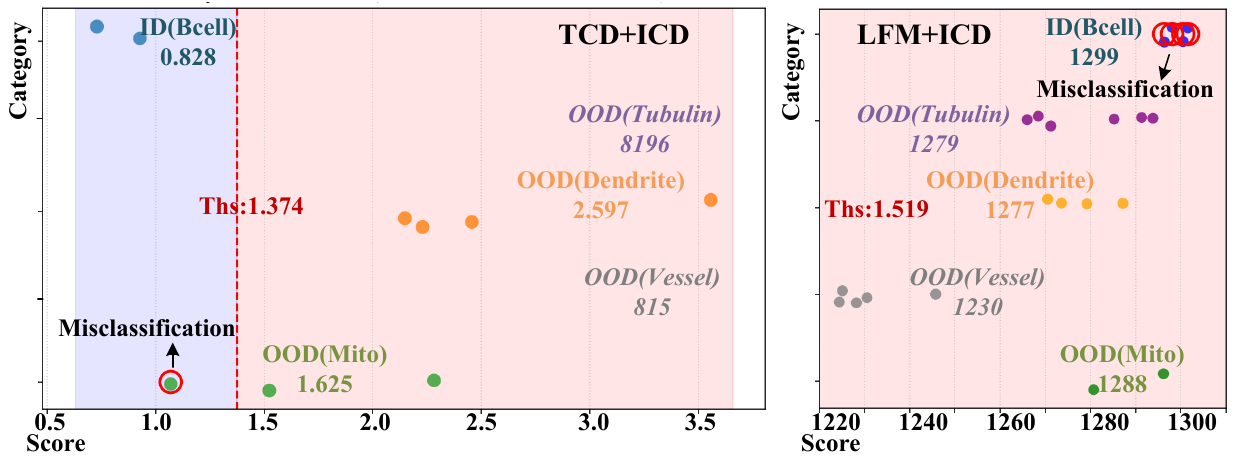}
  \end{subfigure}

  \caption{\textbf{Framework comparison between TCD+ICD (left) and LFMNet+ICD (right).} 
   }
   \label{fig:icd_finetune_combine}
\end{figure}

To further enhance the model’s generalization capability across different biological categories, we evaluated the integrated ICD module. As illustrated in Figure~\ref{fig:icd_finetune_combine}, we designate Bcell as the ID category and construct a mixed test set containing five biological structures to assess the recognition performance of TCD+ICD and LFMNet+ICD. In the figure, the vertical axis denotes the biological category of each test sample, while the horizontal axis represents the Score computed by the ICD module. A lower Score indicates a higher similarity between the test and ID samples, and samples with Scores below a predefined threshold are classified as ID samples.

Experimental results demonstrate that TCD+ICD effectively distinguishes most ID and OOD samples, whereas LFMNet+ICD fails to achieve stable recognition performance, showing difficulty in separating these categories. This finding further verifies the high compatibility and synergy between the ICD module and the proposed TCD framework.

Furthermore, we conducted targeted fine-tuning experiments on the detected OOD samples (see supplementary material for details). The results show that, compared with training from scratch, the ICD-guided fine-tuning process reduces the time required to reach a loss of $10^{-3}$ by approximately half, significantly improving the model’s adaptation efficiency. Notably, as shown in Figure~\ref{fig:icd_finetune_combine}, although a few OOD samples are misclassified as ID samples due to their structural similarity, TCD still maintains robust reconstruction performance and effectively generates valid 3D volumes, benefiting from its superior cross-sample generalization capability.

\section{Limitations and Future Work}

In this study, we proposed a high-efficiency, end-to-end diffusion-based 3D reconstruction method for LFM, termed TCD, which is built upon a lightweight conditional U-Net architecture. Our method achieves high-accuracy 3D imaging while significantly improving inference efficiency, offering a pioneering solution for efficient and high-fidelity LFM reconstruction. However, the current ICD module is not yet widely applicable to other models, and a more generalized reconstruction pipeline has yet to be established.

Although the proposed TCD+ICD framework demonstrates superior performance in LFM 3D reconstruction and exhibits strong cross-sample generalization capability, two directions remain open for further exploration:  
(1) optimizing macroscopic structure reconstruction to enhance adaptability to large-scale biological samples; (2) incorporating physical model constraints, such as PSF modeling or optical prior refinement, during training to further improve reconstruction accuracy and robustness under unsupervised or weakly supervised conditions.

\section*{Acknowledgments}
This work was supported by Hebei Natural Science Foundation (No. F2024203081), National Natural Science Foundation of China (No. 62371006), and Beijing Natural Science Foundation (No. 3242008).